\documentclass[letterpaper]{article} 
\usepackage{aaai2026}  
\usepackage{times}  
\usepackage{helvet}  
\usepackage{courier}  
\usepackage[hyphens]{url}  
\usepackage{amssymb}
\usepackage{graphicx} 
\usepackage{natbib}  
\usepackage{caption} 
\usepackage{algorithm}
\usepackage{algorithmic}

\usepackage{newfloat}
\usepackage{listings}
\DeclareCaptionStyle{ruled}{labelfont=normalfont,labelsep=colon,strut=off} 
\floatstyle{ruled}
\newfloat{listing}{tb}{lst}{}
\floatname{listing}{Listing}
\usepackage{multirow}
\usepackage{amsmath}

\title{PAL: Personal Adaptive Learner}
\author{
Megha Chakraborty, Darssan L. Eswaramoorthi, Madhur Thareja, Het Riteshkumar Shah, Finlay Palmer, Aryaman Bahl, Michelle A Ihetu, Amit Sheth
}
\affiliations{
     Artificial Intelligence Institute, University of South Carolina\\

}

\begin{document}

\maketitle

\begin{abstract}
AI-driven education platforms have made some progress in personalisation, yet most remain constrained to static adaptation—predefined quizzes, uniform pacing, or generic feedback—limiting their ability to respond to learners’ evolving understanding. This shortfall highlights the need for systems that are both context-aware and adaptive in real time.
We introduce PAL (Personal Adaptive Learner), an AI-powered platform that transforms lecture videos into interactive learning experiences. PAL continuously analyzes multimodal lecture content and dynamically engages learners through questions of varying difficulty, adjusting to their responses as the lesson unfolds. At the end of a session, PAL generates a personalized summary that reinforces key concepts while tailoring examples to the learner’s interests. By uniting multimodal content analysis with adaptive decision-making, PAL contributes a novel framework for responsive digital learning. Our work demonstrates how AI can move beyond static personalization toward real-time, individualized support, addressing a core challenge in AI-enabled education.
\end{abstract}

\begin{links}
    \link{Code}{https://tinyurl.com/3c3vx2zn}
    \link{Video}{https://tinyurl.com/yc42pj55}
\end{links}

\section{Introduction}
\label{sec:Introduction}

Artificial intelligence is increasingly used in digital education to improve personalization and engagement. However,ver, most lecture-based platforms remain static: learners watch videos passively, receive uniform quizzes, receive receive limited feedback. Such systems fail to adapt to individual learners in real time, often leaving students either under-challenged or overwhelmed, both of which reduce motivation and retention. Addressing this gap is key for scalable, equitable, and engaging learning environments.

Existing adaptive learning approaches, such as pretests or rigid rule-based adjustments, capture only coarse measures of ability \cite{claned2025, hyperspace2025}. They struggle to keep learners consistently within their optimal “learning zone” where the material is neither too easy nor too difficult.

We present the \textbf{Personal Adaptive Learner (PAL)}, a system that transforms lecture videos into interactive learning sessions. PAL integrates multimodal video analysis with an adaptive question engine that periodically inserts multiple-choice questions during playback. A Hybrid Reinforcement Learning (RL) algorithm dynamically adjusts the difficulty of the question based on the accuracy, streaks, and timing of the responses, ensuring that learners remain engaged while progressively challenged. At the end of each lesson, PAL produces a personalized summary that reinforces key concepts and connects them to the specific interests of the learner.

The contributions of this demo are threefold:
\begin{itemize}
    \item \textbf{PAL system:} A platform that unites lecture content with adaptive, real-time questioning.
    \item \textbf{Hybrid RL algorithm:} An adaptive decision making engine that balances stability with exploration for personalized learning.
    \item \textbf{Live demonstration:} A web-based demo showcasing how PAL sustains engagement and provides tailored post-learn feedback.
\end{itemize}

PAL highlights how AI can move beyond static personalization toward individualized real-time learning support. 

\begin{figure}[H]
    \centering
    \includegraphics[width=0.7\columnwidth]{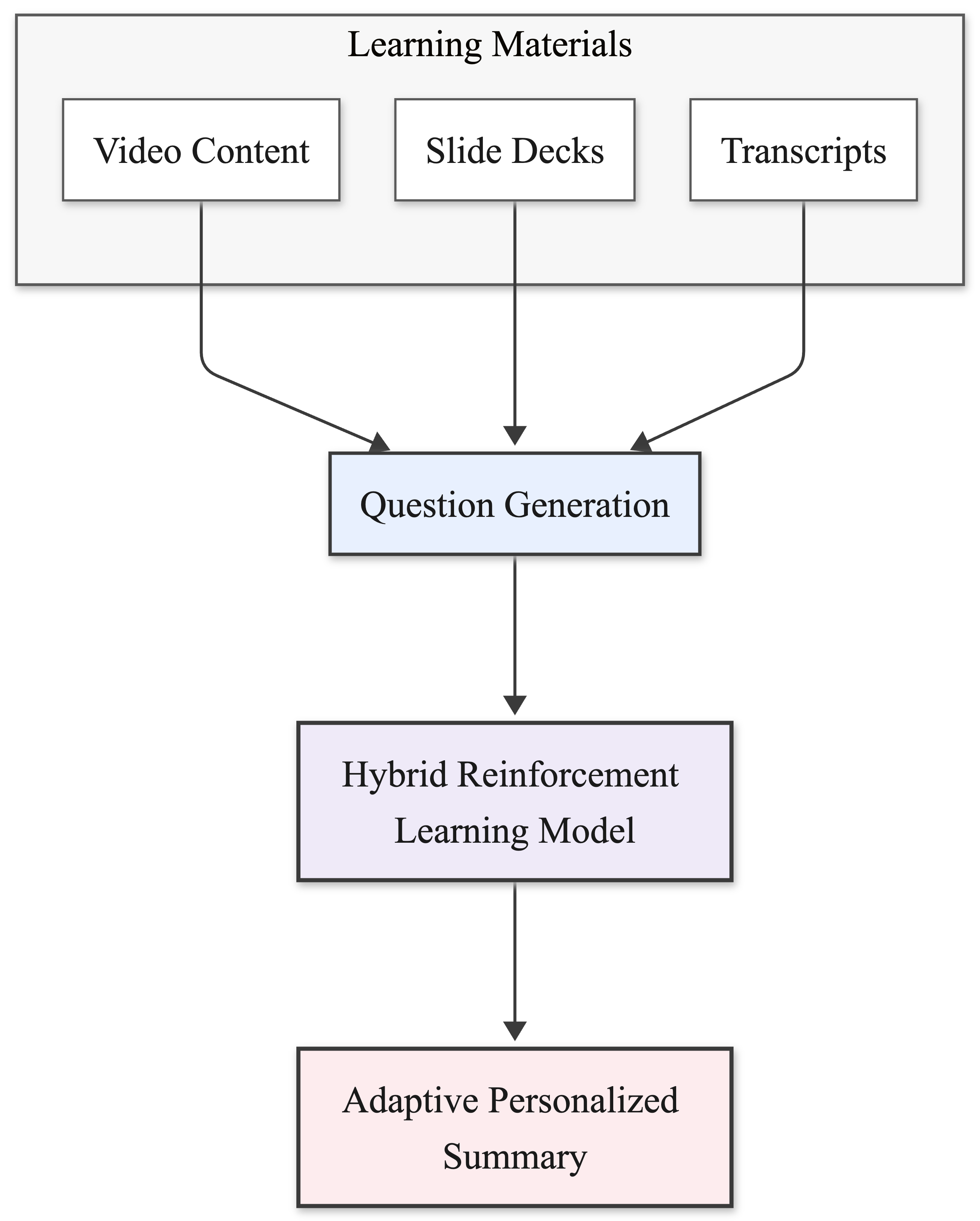}
    \caption{Architecture Diagram of PAL}
    \label{fig:singlecol}
\end{figure}

\section{Video-to-Question Dataset Generation}

Given a lecture video $V$ and transcript $T = \{(t_i, u_i)\}_{i=1}^n$, PAL uses a four-stage pipeline to generate timestamped, difficulty-rated questions \cite{koppolu2025dynamic, singh2025agentic}.  

\textbf{Transcript Analyzer.} Scans $T$ for candidate points via linguistic cues (e.g., ``is defined as'') or every $N$ sentences \cite{garrison2006revisiting}.  

\textbf{Context Validator.} At each point, extracts a frame, applies OCR for slide text, and uses LLaVA-mini for visual description. These combine with transcript text into a context bundle \cite{abinaya2024automated}.  

\textbf{Question Generator.} Produces Q-A pairs from context. Difficulty is heuristic-based: factual (Easy), conceptual (Medium), applied (Hard). Supports MCQs and lightweight LLM generation \cite{du2018harvesting, liu2025thus}.  

\textbf{Difficulty Rater.} Labels difficulty via rules (``what is'' $\rightarrow$ Easy, ``why/how'' $\rightarrow$ Medium, ``apply/predict'' $\rightarrow$ Hard), with LLM fallback.  

\textbf{Master Pipeline.} The integrated \texttt{tri\_plus\_one\_pipeline} outputs structured JSON $(q,a,d,t,c)$ for adaptive learning, quizzes, and study aids.  

This modular design is lightweight, extensible, and ensures contextual accuracy by combining speech, vision, and text.

\section{Adaptive Difficulty via Hybrid RL}

\paragraph{Problem.}
At interaction $t$, the learner state $\mathbf{x}_t\!\in\!\mathbb{R}^6$ encodes
(skill, recent accuracy, normalized response time, streak momentum, learning velocity, confidence) \cite{zheng2024dynamic, liu2025ghpo}.
We choose difficulty $a_t\!\in\!\{\text{Easy},\text{Medium},\text{Hard}\}$ to maximize long‑term reward:
{\small
\[
\pi^\star = \arg\max_{\pi}\;
\mathbb{E}\!\left[\sum_{k=0}^{\infty}\gamma^k\,R_{t+k+1}
\;\middle|\; a_t \sim \pi(\mathbf{x}_t)\right],
\quad \gamma \in [0,1).
\]
}

\paragraph{Reward (composite).}
We use a compact additive shaping:
\[
r_t = r_{\text{acc}} + r_{\text{time}} + r_{\text{prog}} + r_{\text{mom}},
\]
where $r_{\text{acc}}$ is $1$ for a correct response and $-0.5$ for an incorrect one, while 
$r_{\text{time}}\in[0,0.3]$, $r_{\text{prog}}\in[0,0.2]$, and $r_{\text{mom}}\in[0,0.1]$ encourage timely answers, sensible difficulty progression, and streak momentum respectively.


\paragraph{Statistical prior (IRT with stability buffers).}
We impose a smooth prior over difficulties using a two-parameter logistic (2PL) model:
\[
p_{\text{stat}}(d \mid \mathbf{x}_t) \;\propto\; \sigma\!\big(a_d(\theta_t - b_d)\big),
\]
where $a_d$ and $b_d$ are the item discrimination and difficulty parameters, and $\theta_t$ is the learner skill estimate. 
To stabilize adaptation, we apply asymmetric thresholds (promote if recent accuracy $\geq 0.75$, demote only if $\leq 0.35$) 
together with short cooldown and hold periods, which prevent rapid back-and-forth switching between levels \cite{matteucci2012prior, lee2006reliability}.

\paragraph{RL head (bandit + Q‑learning).}
With $\varepsilon$‑greedy selection \cite{vivanti2021can, cannelli2023hedging} and three actions, we update:
\[
Q_{t+1}(a_t)
\leftarrow
Q_t(a_t) + \alpha\!\left[R_t + \gamma \max_{a'} Q_t(a') - Q_t(a_t)\right].
\]

\paragraph{Hybrid stochastic blending.}
We form a confidence‑scaled mixture:
\begin{align}
\pi_t(d\mid \mathbf{x}_t)
&= (1-w_t)\,p_{\text{stat}}(d\mid \mathbf{x}_t) 
   + w_t\,p_{\text{RL}}(d\mid \mathbf{x}_t), \\
w_t &= \min\!\big(w_{\max},\; w_0+\kappa\,c_t\,\mathrm{progress}_t\big).
\end{align}

so the probability of choosing the RL suggestion rises with evidence ($c_t$) and session progress, capped by $w_{\max}$ (e.g., $0.8$). Both heads are updated every step; the statistical prior stabilizes early phases while RL learns a personalized preference.

\section{Personalized Post-Lesson Summary}

At the end of a lecture, PAL generates a personalized summary based on the full learner profile \cite{essa2023personalized}. The summary highlights \textbf{Territory Mastered} and \textbf{Discovery Zone}.

\textbf{Stage 1 - Smart extraction with semantic search.} 
First, the backend preprocesses the whole lecture: it breaks the transcript into sentences and turns each sentence into a numeric vector with a SentenceTransformer \cite{ruas2017keyword}. This creates a “semantic map” of the lecture.
When a student asks about a concept (for example, “Correlation vs Causation”), the system converts that concept into a vector too and does a fast similarity search on the semantic map. That way it finds the most relevant sentences even if they use different words or are spread out. The process ignores conversational noise and collects a compact, context-rich block of raw text to use in the next step.

\textbf{Stage 2 - Context-aware synthesis with Llama 3.2.}
The collected text (which may be fragmented) is sent to our Llama 3.2 1B Instruct model. Instead of a simple “summarize” command, the model receives a carefully designed prompt that makes it act like an expert teaching assistant. The prompt tells the model to use the lecture text as the main source but also allows it to fill gaps and clarify awkward phrasing using its own knowledge. The prompt can also adapt based on the student’s needs (e.g., request more detail for topics the student struggled with). The result is a clear, well-structured explanation that stays faithful to the lecture while improving readability and completeness.

\section{Discussion}
\label{sec:Discussion}

PAL (Personal Adaptive Learner) adapts instruction by generating timestamped, difficulty-rated questions from lecture videos through a multimodal, agentic pipeline. Unlike static AI education tools, PAL integrates transcript analysis, visual context validation, question generation, and difficulty rating to align content with learner performance in real time. The evaluation of PAL will be made available online through an interactive interface, allowing readers to explore example transcripts, generated questions, and difficulty ratings. Future work will expand feedback loops, integrate collaborative learning, and evaluate classroom impact.

\bibliography{references}

\end{document}